\documentclass[letterpaper]{article} 
\usepackage{aaai25}  
\usepackage{times}  
\usepackage{helvet}  
\usepackage{courier}  
\usepackage[hyphens]{url}  
\usepackage{graphicx} 
\usepackage{natbib}  
\usepackage{caption} 
\usepackage{booktabs}       
\usepackage{amsfonts}       
\usepackage{nicefrac}       
\usepackage{microtype}      

\usepackage{tcolorbox}      

\usepackage{subfig}
\usepackage{amsmath}
\usepackage{multirow}

\usepackage{makecell}

\usepackage{algorithm}
\usepackage{algorithmic}

\usepackage{newfloat}
\usepackage{listings}
\DeclareCaptionStyle{ruled}{labelfont=normalfont,labelsep=colon,strut=off} 
\floatstyle{ruled}
\newfloat{listing}{tb}{lst}{}
\floatname{listing}{Listing}
\title{Destroy and Repair Using Hyper-Graphs for Routing}
\author{
    Ke Li\textsuperscript{\rm 1, \rm 2}, Fei Liu\textsuperscript{\rm 2}, Zhengkun Wang\textsuperscript{\rm 1\thanks{Corresponding author}}, Qingfu Zhang\textsuperscript{\rm 2}    
}
\affiliations{
    \textsuperscript{\rm 1}School of System Design and Intelligent Manufacturing, Southern University of Science and Technology\\
    \textsuperscript{\rm 2}Department of Computer Science, City University of Hong Kong\\

    12250110@mail.sustech.edu.cn, 
    fliu36-c@my.cityu.edu.hk, 
    wangzhenkun90@gmail.com,    
    qingfu.zhang@cityu.edu.hk
}

\usepackage{bibentry}

\begin{document}

\maketitle

\begin{abstract}
Recent advancements in Neural Combinatorial Optimization (NCO) have shown promise in solving routing problems like the Traveling Salesman Problem (TSP) and Capacitated Vehicle Routing Problem (CVRP) without handcrafted designs. Research in this domain has explored two primary categories of methods: iterative and non-iterative. While non-iterative methods struggle to generate near-optimal solutions directly, iterative methods simplify the task by learning local search steps. However, existing iterative methods are often limited by restricted neighborhood searches, leading to suboptimal results. To address this limitation, we propose a novel approach that extends the search to larger neighborhoods by learning a destroy-and-repair strategy. Specifically, we introduce a Destroy-and-Repair framework based on Hyper-Graphs (DRHG). This framework reduces consecutive intact edges to hyper-edges, allowing the model to pay more attention to the destroyed part and decrease the complexity of encoding all nodes. Experiments demonstrate that DRHG achieves state-of-the-art performance on TSP with up to 10,000 nodes and shows strong generalization to real-world TSPLib and CVRPLib problems. 
\end{abstract}

\begin{links}
\link{Code}{https://github.com/CIAM-Group/DRHG}
\end{links}

\begin{figure*}[htbp]
\centering
\includegraphics[width=0.8\textwidth]{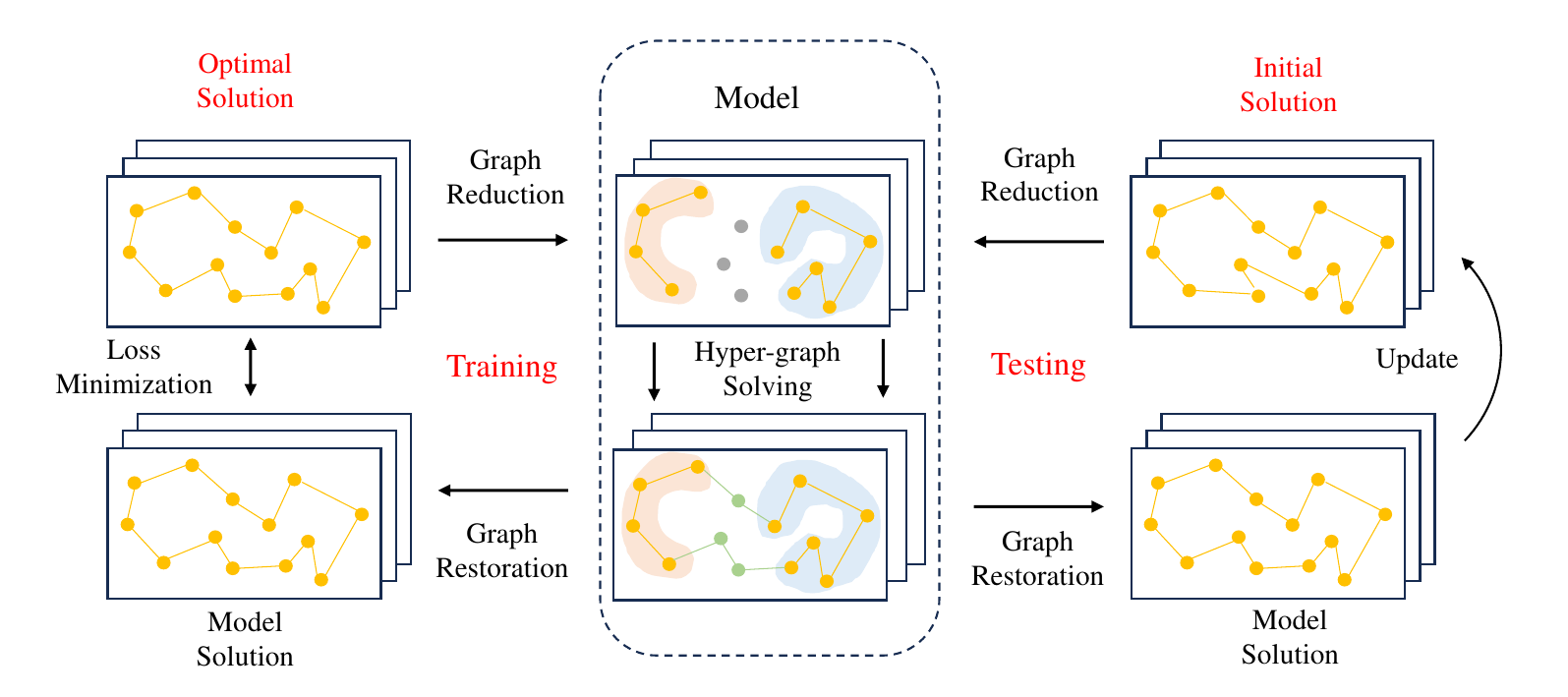} 
\caption{Pipeline of Destroy-and-Repair using Hyper-Graphs   
}
\label{pipeline}

\end{figure*}

\section{Introduction}

Routing problems are significant combinatorial optimization problems with broad real-world applications in logistics, transportation, and manufacturing. Their NP-hard nature poses a significant challenge to the application of exact methods. Heuristics sacrifice the optimality while can obtain near-optimal solutions in a reasonable time. However, the development of heuristics usually relies on human designs with domain expert knowledge, which hinders their practical applications. 

Neural Combinatorial Optimization (NCO), which trains a neural network to learn heuristics to solve routing problems without handcraft design, has gained much attention. The existing NCO methods can be roughly classified into two categories: 1) non-iterative and 2) iterative methods. 

In non-iterative methods, the neural solvers construct a solution in one shot \cite{vinyals2015pointer, kool2018attention, kwon2020pomo}. Most of these works train neural solvers to determine the next node in an auto-regressive manner, i.e., nodes are selected one by one to be added to the end of a partial solution. Others learn to predict a heuristic, such as a heatmap, and then construct a solution based on the learned information. These works can generate reasonable solutions in a short time. Nevertheless, these non-iterative methods may lead to irreversible consequences if an error occurs in one of the construction steps, thus placing excessive demands on the model's capability to narrow the optimality gap for large-scale problems.

Iterative methods adopt neural solvers to tackle a subproblem in each iteration rather than solve the entire problem at once. The iterative approach reduces the burden of neural solvers and increases the performance, leading to state-of-the-art results. Some existing iterative NCO methods \cite{d2020learning2opt,wu2021learning,ma2021DACT} primarily focus on learning low-level operators within small neighborhoods, such as k-opt or swap. Others follow a destroy-and-repair manner, iteratively destroying the solution into a partial solution and then reconstructing the destroyed nodes, operating within a large neighborhood and excelling at producing high-quality solutions.

However, the neural networks are trained in a conventional way, either using Reinforcement Learning (RL) \cite{LCP,cheng2023select,ye2024glop, zheng2024udc} or Supervised Learning (SL) \cite{luo2023lehd,luo2024SIL} without being tailored for the destroy and repair framework. Therefore, they can only deal with the destruction of one segment, bringing challenges in reducing the optimality gap.

To address the issue, we propose a novel iterative NCO method for routing, termed Destroy and Repair by Hyper-Graphs (DRHG). We employ SL to train a model that approaches the best repair after the destruction. Specifically, after the destruction, the complete tour becomes some segments of consecutive edges and some isolated nodes. We reduce the segments to hyper-edges to build a hyper-graph, then fix them during the repair. The model learns to connect isolated nodes and fixed hyper-edges to form a reduced solution, which is restored to a complete solution later. Thanks to the condensed formulation of the hyper-graph, the scale of model input depends only on the degree of the destruction but not the scale of the original problem. This ensures our method has a low computational complexity and allows our method to iterate on large-scale problems. 

Our contributions can be summarized as follows:

\begin{itemize}
    \item We propose a novel NCO framework of destroy-and-repair for routing problems. By learning to repair a destroyed problem in a supervised way, our model can search in large neighborhoods more efficiently.
    \item We adopt a condensed hyper-graph formulation of the destroyed problem by reducing consecutive edges to fixed hyper-edges, which decreases the computational complexity and enables the model to iterate on large-scale problems.
    \item The experiments show that our method achieves state-of-the-art performance on TSPs from 100 nodes to 10K nodes, and also competitive results on CVRP. Our method generalizes well to real-world instances as well. 
\end{itemize}

\section{Related Works}

\subsection{Non-iterative NCO Routing Solvers}

\subsubsection{One-shot Constructive Solvers}
One-shot constructive methods are one of the earliest lines of work that use NCO to solve routing problems. Pioneering works \cite{vinyals2015pointer, bello2016neural, nazari2018reinforcement} show that neural networks such as RNN can be trained to solve routing problems. Inspired by \citet{vaswani2017attention}, some works
\cite{kool2018attention, deudon2018learning} introduce the Transformer architecture to build more powerful NCO models and achieve promising performance. Following their works, various Transformer-based methods \cite{kwon2020pomo, drakulic2023bq, luo2023lehd} emerged. Although they have made progress in training methods or model structures, the one-shot approach can hardly further narrow the performance gap to the optimal results. 

\subsubsection{Heatmap-based Solvers}
Heatmap-based methods aim to predict an informative heatmap to expedite the search process and enhance the quality of solutions. \citet{joshi2019efficient} train a Graph Neural Network (GNN) in SL to predict the probabilities of edges to be optimal, then use the beam search to generate feasible solutions. \citet{kool2022deep} adopt dynamic programming and eliminate dominated partial solutions to reduce searching time. The most prominent works \cite{fu2021generalize, sun2023difusco} in this category employ Monte Carlo Tree Search (MCTS) to construct solutions. Leveraging MCTS reduces the stringent requirements for the accuracy of edge score predictions. However, most heatmap-based methods are limited to TSPs, as their search strategies are incompatible with problems involving additional constraints, such as CVRPs.

\subsection{Iterative NCO Routing Solvers}

Most existing iterative NCO routing solvers focus on learning low-level operators searching within small neighborhoods. \citet{chen2019neural_rewriter} employ a region-picking policy to identify a node for relocation and a rule-picking policy to determine the target position for the node's movement. \citet{d2020learning2opt, sui2021learning3opt} propose to learn 2-opt or 3-opt steps to improve the solution. Furthermore, \citet{lu2019learning, wu2021learning} utilize a pool of operators from which the model selects, demonstrating superior performance compared to approaches that rely on a single operator. \citet{ma2021DACT} propose a Dual-Aspect Collaborative Transformer (DACT) with a Cyclic Positional Encoding (CPE) method and a Dual-Aspect Collaborative Attention (DAC-Att) to encode problems, which achieves pretty good performance. However, iterative NCOs with low-level operators are limited to solving small-size problems due to the extensive number of iterations required for convergence. Moreover, the overall quality of local optimal of small neighborhoods is inferior, which implies that the final solutions obtained by these methods are often sub-optimal.

Other iterative NCO routing solvers focus on reconstructing a partial solution of node sequence. Either trained with RL \cite{LCP, cheng2023select, ye2024glop} or SL \cite{luo2023lehd, luo2024SIL}, the models learn to reconstruct a segment given the starting node and the ending node. By operating within a large neighborhood, these methods outperform those using low-level operators. However, the neighborhoods that these methods can search in are still limited since the nodes outside the segment remain unaltered. Therefore, two nodes that are spatially close but far away in the solution may have no chance of being reconnected together. In contrast, our framework enables a more flexible neighborhood search by permitting arbitrary destruction and, subsequently, the repair of reconnecting the segments with isolated nodes.

\section{Methodology}

\subsection{DRHG Framework}

Schematically illustrated in Fig. \ref{pipeline}, we reformulate our destroy-and-repair approach as a graph reduction, hyper-graph solving, and graph restoration process. For a graph representing the incomplete solution where a set of edges is destroyed, we reduce the graph by encoding the remaining consecutive edges as hyper-edges. As these edges remain unchanged during the repair, redefining them as fixed hyper-edges helps reduce the complexity of the problem for the model. Then, we train the model in a supervised way to solve the reduced problem on the hyper-graph. In the testing phase, we iteratively destroy the current solution to obtain a hyper-graph, solve the resulting hyper-graph, and recover the hyper-graph solution on that of the original problem.

\subsection{Hyper-graph Representation}\label{sec: represent hyper-graph}

Mathematically, a hyper-graph is a special graph where an edge can join any number of vertices. Formally, a hyper-graph is defined as $\mathcal{G}=(\mathcal{V},\mathcal{E})$, in which $\mathcal{V}$ is the vertex (node) set and $\mathcal{E}$ is the hyper-edge set. Using hypergraph neural networks for embedding hypergraphs is intuitive, but challenging. Specifically, when constructing a solution sequentially, it becomes necessary to align the embeddings of nodes and edges in order to predict the subsequent node or hyper-edge, which may be hard for models. Even if we train an excellent model to predict the sequence, resolving the solution with respect to an undirected hyper-graph remains a non-trivial challenge. Since each hyper-edge has two possible directions, resolving the best complete solution may require a huge number of enumerations. Therefore, we propose to use two endpoints to represent a hyper-edge.

Note a TSP instance of $n$ nodes by the node coordinates as $V=\{(x_1, y_1), \ (x_2, y_2),\  ..., \ (x_n, y_n)\}$. After the destruction, $h$ undestroyed segments constitute the directed hyper-edges set of size $2h$, i.e., $E=\{e^i=(i_1, i_2, ..., i_{p_i})\ | i=1,2,...,2h\}$, where $p_i$ is the number of nodes in the directed hyper-edge $e^i$.

For hyper-graph reduction, we remove the middle nodes inside the hyper-edges and keep only the endpoints to represent the hyper-edge, i.e., $e^i=(i_1, i_p)$. Consequently, in the reduced graph, we have one set of isolated nodes $A$, one set of endpoint nodes $B$, and one set of reduced hyper-edges $E_r$. The hyper-graph size is $m=|A|+|B|$. Then, the input feature of the reduced graph is formulated as:

\begin{equation}
  r_i =(x_i^a, y_i^a, x_i^b, y_i^b, flag_i), i=1,2,...,m,
\end{equation}
    
\begin{equation}
    (x_i^a, y_i^a) = (x_i, y_i),
\end{equation}

\begin{equation}
(x_i^b, y_i^b) = \begin{cases}
	      (x_i, y_i), & if\ i\in A,\\
	      (x_j, y_j), & if\ i\in B\ and (i,j) \in E_r,
		   \end{cases}
\end{equation}
where $r_i$ is the input feature for the model, and $flag_i$ is a binary variable to indicate whether a node is an endpoint node or an isolated node.

Similarly, for a CVRP instance of $n$ customers and a depot noted as $0$, we can define the problem by the node coordinates and the demands: $V=\{(x_0, y_0, 0), \ (x_1, y_1, d_1),\  ..., \ (x_n, y_n, d_n)\}$,  where $d_i$ is the demand of node $i$. To simplify the problem, we destroyed all edges connected to the depot. Then, the input feature of the reduced graph for CVRP can be formulated as follows:

\begin{equation}
  r_i =(x_i^a, y_i^a, x_i^b, y_i^b, flag_i, dr_i), i=1,2,...,m,
\end{equation}

\begin{equation}
dr_i = \begin{cases}
	      d_i, & if\ i\in A,\\
	      \sum_{k}d_k, & if\ i\in B\ and\ k\in (i_1, i_2, ..., i_{p_i}).
		   \end{cases}
\end{equation}

\begin{figure}[htbp]
\centering
\includegraphics[width=0.5\textwidth]{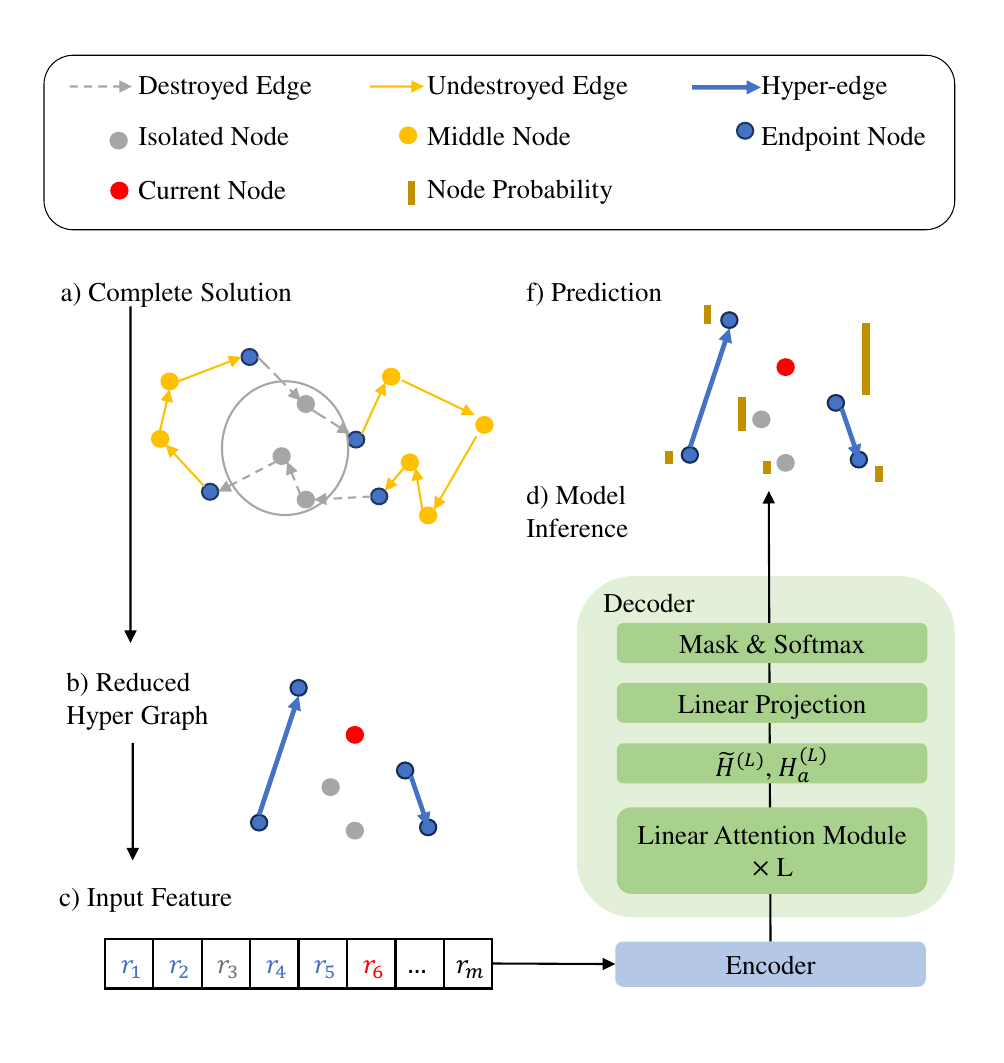} 
\caption{Model structure of DRHG}
\label{Model structure}
\end{figure}

\subsection{Model Structure}\label{sec: model}
As shown in Fig. \ref{Model structure}, given the input features of the reduced problems, our model yields a prediction of the next node through a light encoder and a heavy decoder.

\paragraph{Encoder} The encoder consists of a single linear projection layer, which transforms the input $r_i \in \mathbb{R}^{d_i}$ into embedding $h_i^{(0)} \in \mathbb{R}^{d_h}$.

\paragraph{Decoder} The decoder has a slightly changed linear attention module in \citet{luo2024SIL}. At each step $t$, the decoder takes the node embeddings of the first node $h_f^{(0)}$, the current node $h_c^{(0)}$, and the remaining unselected nodes $H_a^{(0)} = \{h_i^{(0)} | i = 1,2, ..., m-t\}$ as inputs. Then, the first node $h_f^{(0)}$ and the current node $h_c^{(0)}$ are used to generate $r$ virtual representative nodes embeddings $\Tilde{H}^{(0)} = \{h_j^{(0)} | j=1,2, ...,r\}$, which combined with $H_a^{(0)}$, form the input of the first linear attention module.

Then, we stack $L$ linear modules as the main component of the decoder. A linear attention module is composed of an aggregating attention layer and a broadcasting attention layer. The aggregating layer aggregates information to the representative nodes, and then the broadcasting layer broadcasts gathered information to all nodes in the graph. The details of the linear attention module are provided in Appendix \ref{appendix-model}. Note the $l$-th linear attention module as $L-Att^{(l)}$, we have
\begin{equation}
    \Tilde{H}^{(l)}, H_a^{(l)} = L-Att^{(l)}(\Tilde{H}^{(l-1)}, H_a^{(l-1)}).
\end{equation}

After $L$ attention module, we obtain a hidden representation $\Tilde{H}^{(L)}$ and and $H_a^{(L)}$. Then we take only $H_a^{(L)}$ to calculate the probability of selecting the next node by a linear projection layer and the softmax function:

\begin{equation}
a_i=\phi(h_i^{(L)}W_o),
\end{equation}

\begin{equation}
p_i=\frac{e^{a_i}}{\Sigma_j^{e^{a_j}}}.
\end{equation}

\subsection{Training Scheme}\label{sec: model}

We use SL to train our model. We apply the clustering destruction, as optimal edges are more likely to connect proximal nodes. Furthermore, the distributions of reduced problems after clustering destructions are more consistent across problems of different scales. We adopt the coordinate transformation in \citet{ye2024glop} to enhance the distribution homogeneity and consistency. For hyper-edges, once one endpoint is selected, the subsequent node must be the other endpoint. This behavior is dictated by the constraint rather than the model. Correspondingly, we introduce a masking mechanism to block the associated gradients. Additionally, destroying the problem by k-nearest neighbors results in a variable number of segments and makes the hyper-graph size differ across instances. This variability introduces instability during the training process. To tackle this problem, we design a special destruction scheme to get fixed-size hyper-graphs. We detail this method in Appendix \ref{appendix-alignement}.

\section{Experiments}
We compare our method with other representative learning-based and classical solvers on
TSP and CVRP instances with different scales and the instances in the real world.
\subsection{Experiment Setup}

\subsubsection{Implementing Details}
We set the embedding dimension of the encoder to 128. The decoder is composed of 6 linear attention modules, and each has 8 attention heads and 16 representative starting nodes. The hidden dimension of the feed-forward layer is set to 512. 

For TSP, we train the model for 100 epochs on 1,000,000 TSP100 instances. We fine-tune 20 epochs on 10,000 TSP1000 instances for large-scale problems. For CVRP, we train the model for 100 epochs on 1,000,000 CVRP100 instances. We use a batch size of 1024 and sample the training sample size in $[20, 0.8n]$ where $n$ is the problem size. As the fixed-size destruction scheme will discard a small part of the samples, the true batch size is around 800. We use the cross-entropy loss and the Adam optimizer \cite{Adam}. The initial learning rate is 1e-4, and the decay rate is 0.97 per epoch. We train and test our model with a single NVIDIA GeForce RTX 3090 GPU with 24GB memory.

\subsubsection{Baselines}
We compare our method with:

\textbf{1) Classical Solvers:} Concorde \cite{applegate2006concorde}, LKH3 \cite{LKH3}, and HGS \cite{HGS}; 

\textbf{2) Traditional Heuristic:} Random Insertion, Sweep; 

\textbf{3) Construction-based Method:} POMO \cite{kwon2020pomo}, BQ \cite{drakulic2023bq};

\textbf{4) Heatmap-based Method:} Att-GCN+MCTS \cite{fu2021generalize}, DIMES \cite{qiu2022DIMES}, and DIFUSCO \cite{sun2023difusco};

\textbf{5) Segment-reconstruction Method:} LEHD \cite{luo2023lehd}, GLOP \cite{ye2024glop} and SIL \cite{luo2024SIL}; 

\textbf{6) Operator-iteration Method:} Neural Rewriter \cite{chen2019neural_rewriter}, Learning 2-Opt \cite{d2020learning2opt}, Learning 3-Opt \cite{sui2021learning3opt}, and DACT \cite{ma2021DACT}.

For most baseline methods, we run their source code with default settings. The result of Att-GCN+MCTS \cite{fu2021generalize}, DIMES \cite{qiu2022DIMES}, and DIFUSCO \cite{sun2023difusco}, SIL \cite{luo2024SIL}, Neural Rewriter \cite{chen2019neural_rewriter}, and Learning 3-Opt \cite{sui2021learning3opt} are taken from their original papers.

\subsubsection{Metrics} 
 
We use the average objective value (Obj.) and optimality gap (Gap) to evaluate the model performance and the inference time (Time) to evaluate the model efficiency. The ground truth labels of TSP are generated by Concorde for TSP100 to TSP1000 and by LKH for TSP above 1000. The ground truth labels of CVRP are generated by HGS.

\subsubsection{Testing}
We test our method on TSPs from 100 to 10,000 and CVRPs from 100 to 1000. There are 10,000 instances for TSP100 and CVRP100, 128 instances for problems of size 200 to 5,000, and 16 instances for TSP10,000. We use random insertion to generate initial solutions for TSP and sweep for CVRP. Regarding to the k-nn destruction, we sample $k\in [20, min(1000, n)]$ for TSP and $k\in [20, min(200, n)]$ for CVRP, where $n$ is the problem scale. For simplicity, we disconnect all nodes adjacent to the depot in CVRP. We set the number of iterations to 1000.   

\begin{table*}[htbp]
  \centering
  \renewcommand{\arraystretch}{1.4}
  \renewcommand{\tabcolsep}{3pt} 
    \begin{tabular}{l|ccc|ccc|ccc}
    \toprule
          & \multicolumn{1}{c}{} & \multicolumn{1}{c}{\textbf{TSP100}} & \multicolumn{1}{c|}{} 
          & \multicolumn{1}{c}{} & \multicolumn{1}{c}{\textbf{TSP200}} & \multicolumn{1}{c|}{} 
          & \multicolumn{1}{c}{} & \multicolumn{1}{c}{\textbf{TSP500}} & \multicolumn{1}{c}{} \\
          
    \multicolumn{1}{c|}{\textbf{Method}} & \multicolumn{1}{c}{\textbf{Obj.}} & \multicolumn{1}{c}{\textbf{Gap}} & \textbf{Time} & \multicolumn{1}{c}{\textbf{Obj.}} & \multicolumn{1}{c}{\textbf{Gap}} & \textbf{Time} & \multicolumn{1}{c}{\textbf{Obj.}} & \multicolumn{1}{c}{\textbf{Gap}} & \textbf{Time} \\
    \midrule
    LKH3  & \multicolumn{1}{c}{7.763 } & \multicolumn{1}{c}{0.000\%} & 0.34s & \multicolumn{1}{c}{10.704 } & \multicolumn{1}{c}{0.000\%} & 1.88s & \multicolumn{1}{c}{16.522 } & \multicolumn{1}{c}{0.000\%} & 15.0s \\
    Concorde & \multicolumn{1}{c}{7.763 } & \multicolumn{1}{c}{0.000\%} & 0.20s & \multicolumn{1}{c}{10.704 } & \multicolumn{1}{c}{0.000\%} & 1.41s & \multicolumn{1}{c}{16.522 } & \multicolumn{1}{c}{0.000\%} & 15.0s \\
    Random Insertion & \multicolumn{1}{c}{8.513 } & \multicolumn{1}{c}{9.662\%} & 0.00s & \multicolumn{1}{c}{11.948 } & \multicolumn{1}{c}{11.627\%} & \textless0.01s & \multicolumn{1}{c}{18.546 } & \multicolumn{1}{c}{12.252\%} & \textless0.1s \\
    \midrule
    Att-GCN+MCTS*  & \multicolumn{1}{c}{7.764 } & \multicolumn{1}{c}{0.037\%} & 0.09s & \multicolumn{1}{c}{10.814 } & \multicolumn{1}{c}{0.884\%} & 0.94s & \multicolumn{1}{c}{16.966 } & \multicolumn{1}{c}{2.537\%} & 2.8s \\
    DIMES* & \multicolumn{1}{c}{-} & \multicolumn{1}{c}{-} & -     & \multicolumn{1}{c}{-} & \multicolumn{1}{c}{-} & -     & \multicolumn{1}{c}{16.840 } & \multicolumn{1}{c}{1.760\%} & 60.5s \\
    DIFUSCO* & \multicolumn{1}{c}{7.780 } & \multicolumn{1}{c}{0.240\%} & -     & \multicolumn{1}{c}{-} & \multicolumn{1}{c}{-} & -     & \multicolumn{1}{c}{16.800 } & \multicolumn{1}{c}{1.490\%} & 1.7s \\
    \midrule
    
    POMO augx8 & \multicolumn{1}{c}{\makecell{7.774\\(±0.231)}} 
    & \multicolumn{1}{c}{\makecell{0.134\%\\(±0.224\%)}} & 0.01s    
    & \multicolumn{1}{c}{\makecell{10.868\\(±0.225)}} 
    & \multicolumn{1}{c}{\makecell{1.534\%\\(±0.523\%)}} & 0.04s 
    & \multicolumn{1}{c}{\makecell{20.187\\(±0.251)}} 
    & \multicolumn{1}{c}{\makecell{22.187\%\\(±0.997\%)}} & 0.5s \\

    BQ bs16 & \multicolumn{1}{c}{\makecell{7.764\\(±0.229)}} 
    & \multicolumn{1}{c}{\makecell{0.015\%\\(±0.057\%)}} & 0.17s      
    & \multicolumn{1}{c}{\makecell{10.717\\(±0.208)}} 
    & \multicolumn{1}{c}{\makecell{0.129\%\\(±0.149\%)}} & 0.94s  
    & \multicolumn{1}{c}{\makecell{16.617\\(±0.212)}} 
    & \multicolumn{1}{c}{\makecell{0.579\%\\(±0.239\%)}} & 5.5s  \\
    \midrule
    
    GLOP (more revision) & \multicolumn{1}{c}{\makecell{7.767\\(±0.234)}}
    & \multicolumn{1}{c}{\makecell{0.046\%\\(±0.126\%)}} & 0.79s 
    & \multicolumn{1}{c}{\makecell{10.774\\(±0.213)}} 
    & \multicolumn{1}{c}{\makecell{0.653\%\\(±0.410\%)}} & 0.33s  
    & \multicolumn{1}{c}{\makecell{16.883\\(±0.214)}} 
    & \multicolumn{1}{c}{\makecell{2.186\%\\(±0.474\%)}} & 0.8s  \\

    LEHD RRC1000 & \multicolumn{1}{c}{\makecell{7.763\\(±0.229)}}
    & \multicolumn{1}{c}{\makecell{0.002\%\\(±0.014\%)}} & 1.04s 
    & \multicolumn{1}{c}{\makecell{10.706\\(±0.206)}}
    & \multicolumn{1}{c}{\makecell{0.0182\%\\(±0.054\%)}} & 4.92s
    & \multicolumn{1}{c}{\makecell{16.550\\(±0.209)}}
    & \multicolumn{1}{c}{\makecell{0.167\%\\(±0.128\%)}} & 33.8s \\
    \midrule
    
    Learning 2-Opt (T=1000) & \multicolumn{1}{c}{7.853} & \multicolumn{1}{c}{1.150\%} & 0.09s & \multicolumn{1}{c}{11.107} & \multicolumn{1}{c}{3.765\%} & 0.20s & \multicolumn{1}{c}{21.339} & \multicolumn{1}{c}{29.158\%} & 0.5s \\
    
    Learning 3-Opt (T=1000)* & \multicolumn{1}{c}{7.850 } & \multicolumn{1}{c}{1.060\%} & 0.23s & \multicolumn{1}{c}{-} & \multicolumn{1}{c}{-} &       & \multicolumn{1}{c}{-} & \multicolumn{1}{c}{-} & - \\
    
    DACT (T=1000) & \multicolumn{1}{c}{7.892 } & \multicolumn{1}{c}{1.653\%} & 0.07s & \multicolumn{1}{c}{12.870 } & \multicolumn{1}{c}{20.252\%} & 0.41s & \multicolumn{1}{c}{20.846 } & \multicolumn{1}{c}{26.171\%} & 1.6s \\
    \midrule

    DRHG (T=1000) & \multicolumn{1}{c}{\makecell{\textbf{7.763}\\\textbf{(±0.229)}}}
    &\multicolumn{1}{c}{\makecell{\textbf{0.000\%}\\\textbf{(±0.007\%)}}}
    & 2.73s      
    & \multicolumn{1}{c}{\makecell{\textbf{10.705}\\\textbf{(±0.206)}}} 
    & \multicolumn{1}{c}{\makecell{\textbf{0.010\%}\\\textbf{(±0.036\%)}}}
    & 9.05s 
    & \multicolumn{1}{c}{\makecell{\textbf{16.540}\\ \textbf{(±0.211)}}} 
    & \multicolumn{1}{c}{\makecell{\textbf{0.111\%}\\ \textbf{(±0.090\%)}}} & 20.6s \\
    \bottomrule

    & \multicolumn{1}{c}{} & \multicolumn{1}{c}{\textbf{TSP1K}} & \multicolumn{1}{c|}{} 
    & \multicolumn{1}{c}{} & \multicolumn{1}{c}{\textbf{TSP5K}} & \multicolumn{1}{c|}{} 
    & \multicolumn{1}{c}{} & \multicolumn{1}{c}{\textbf{TSP10K}} & \multicolumn{1}{c}{} \\
    
     \multicolumn{1}{c|}{\textbf{Method}} & \textbf{Obj.} & \textbf{Gap}   & \textbf{Time}  &  \textbf{Obj.}  & \textbf{Gap}   & \textbf{Time}  &  \textbf{Obj.}  & \textbf{Gap}   & \textbf{Time} \\
    \midrule
    
    LKH3  & 23.12  & 0.00\% & 1.7m  & 50.97  & 0.00\% & 12m   & 71.78  & 0.00\% & 33m \\
    Concorde & 23.12  & 0.00\% & 1m    & 50.95  & -0.05\% & 31m   & 72.00  & 0.15\% & 1.4h \\
    Random Insertion & 26.11  & 12.90\% & \textless1s   & 58.06  & 13.90\% & \textless1s   & 81.82  & 13.90\% & \textless1s \\
    \midrule
    
    Att-GCN+MCTS*  & 23.86  & 3.20\% & 6s    & -     & -     & -     & 74.93  & 4.39\% & 6.6m \\
    DIMES* & 23.69  & 2.46\% & 2.2m  & -     & -     & -     & 74.06  & 3.19\% & 3m \\
    DIFUSCO* & 23.39  & 1.17\% & 11.5s & -     & -     & -     & 73.62  & 2.58\% & 3.0m \\
    \midrule
    
    POMO augx8 & 32.51  & 40.60\% & 4.1s  & 87.72  & 72.10\% & 8.6m  &       & OOM   &  \\

    BQ bs16 & \multicolumn{1}{c}{\makecell{23.43\\(±0.221)}} & \multicolumn{1}{c}{\makecell{1.37\%\\(±0.284\%)}} & 13s  
    & \multicolumn{1}{c}{\makecell{58.27\\(±0.951)}} 
    & \multicolumn{1}{c}{\makecell{10.70\%\\(±1.827\%)}} & 24s   
    & \multicolumn{1}{c}{} & \multicolumn{1}{c}{OOM} &    \\

    GLOP (more revision) & \multicolumn{1}{c}{\makecell{23.78\\(±0.218)}} 
    & \multicolumn{1}{c}{\makecell{2.85\%\\(±0.401\%)}} & 10.2s 
    & \multicolumn{1}{c}{\makecell{53.15\\(±0.231)}} 
    & \multicolumn{1}{c}{\makecell{4.26\%\\(±0.289\%)}} & 1.0m  
    & \multicolumn{1}{c}{\makecell{75.04\\(±0.215)}} & \multicolumn{1}{c}{\makecell{4.39\%\\(±0.153\%)}} & 1.9m  \\
    
    LEHD RRC1000 & \multicolumn{1}{c}{\makecell{23.29\\(±0.220)}} 
    & \multicolumn{1}{c}{\makecell{0.72\%\\(±0.176\%)}} & 3.3m 
    & \multicolumn{1}{c}{\makecell{54.43\\(±0.394)}} 
    & \multicolumn{1}{c}{\makecell{6.79\%\\(±0.671\%)}} & 8.6m  
    & \multicolumn{1}{c}{\makecell{80.90\\(±0.532)}} & \multicolumn{1}{c}{\makecell{12.50\%\\(±0.663\%)}} & 18.6m \\
    
    SIL PRC1000* & 23.31  & 0.82\% & 1.2m  & 51.91  & 1.84\% & 7.6m  & 73.38  & 2.23\% & 13.7m \\
    \midrule
    Learning 2-Opt (T=1000) & 61.15  & 164.50\% & 1.3s  & -     & -     & -     & -     & -     & - \\
    DACT (T=1000) & 29.03  & 25.56\% & 7.8s  &       & OOM   &       &       & OOM   &  \\
    \midrule
    
    DRHG (T=1000) & 23.22  & 0.45\% & 1.72m & 51.98  & 2.05\% & 1.79m & 74.38  & 3.46\% & 3.63m \\

    DRHG-FT (T=1000) 
    & \multicolumn{1}{c}{\makecell{\textbf{23.19}\\\textbf{(±0.210)}}} 
    & \multicolumn{1}{c}{\makecell{\textbf{0.29\%}\\\textbf{(±0.108\%)}}} 
    & 1.66m 
    & \multicolumn{1}{c}{\makecell{\textbf{51.39}\\\textbf{(±0.187)}}} & 
    \multicolumn{1}{c}{\makecell{\textbf{0.88\%}\\\textbf{(±0.087\%)}}} & 1.82m 
    & \multicolumn{1}{c}{\makecell{\textbf{72.85}\\\textbf{(±0.217)}}} & \multicolumn{1}{c}{\makecell{\textbf{1.33\%}\\\textbf{(±0.084\%)}}} & 3.70m \\

    \bottomrule
    
    \end{tabular}%
    \caption{Results on TSP}
  \label{table-tsp-all}%
\end{table*}%

\begin{table*}[htbp]
  \centering
\begin{tabular}{c|c c|c c|c c}
\toprule
\multicolumn{1}{c|}{\multirow{2}[4]{*}{\textbf{Competitors}}} & \multicolumn{2}{c|}{\textbf{TSP100}} & \multicolumn{2}{c|}{\textbf{TSP200}} & \multicolumn{2}{c}{\textbf{TSP500}} \\
\cmidrule{2-7}      & \textbf{Competitor's} & \textbf{Ours}  & \textbf{Competitor's} & \textbf{Ours}  & \textbf{Competitor's} & \textbf{Ours} \\
\midrule
POMO augx8 & 0.134\% & 1.311\% (-) & 1.534\% & 1.269\%(+) & 22.187\% & 1.113\%(+) \\
\midrule
BQ bs16 & 0.015\% & 0.025\% (-) & 0.129\% & 0.076\%(+) & 0.579\% & 0.322\%(+) \\
\midrule
GLOP (more revision) & 0.046\% & 0.004\% (+) & 0.653\% & 0.206\%(+) & 2.186\% & 0.879\%(+) \\
\midrule
LEHD RRC1000 & 0.002\% & 0.003\% (-) & 0.018\% & 0.016\%(+) & 0.167\% & 0.111\%(+) \\
\bottomrule
\end{tabular}%
\caption{Comparison of different methods on TSPs with scale $\textless$ 1,000 given the same running time}
  \label{comparason}%
\end{table*}%


\begin{table*}[htbp]
  \centering
    \renewcommand{\tabcolsep}{10pt} 
    \begin{tabular}{l|p{10em}p{6em}c|p{6em}p{6em}c}
    
    \toprule    
    & \multicolumn{1}{c}{} & \multicolumn{1}{c}{\textbf{CVRP100}} & \multicolumn{1}{c|}{} 
    & \multicolumn{1}{c}{} & \multicolumn{1}{c}{\textbf{CVRP200}} & \multicolumn{1}{c}{} \\
    
    \textbf{Method} & \multicolumn{1}{c}{\textbf{ Obj. }} & \multicolumn{1}{c}{\textbf{Gap}} & \textbf{Time} & \multicolumn{1}{c}{\textbf{ Obj. }} & \multicolumn{1}{c}{\textbf{Gap}} & \multicolumn{1}{c}{\textbf{Time}}  \\
    \midrule
    
    LKH3  & \multicolumn{1}{c}{15.647} & \multicolumn{1}{c}{0.00\%} & 4.32s &      \multicolumn{1}{c}{20.173} & \multicolumn{1}{c}{0.00\%} & 59.06s \\
    HGS   & \multicolumn{1}{c}{15.563} & \multicolumn{1}{c}{-0.53\%} & 1.62s &  \multicolumn{1}{c}{19.946} & \multicolumn{1}{c}{-1.13\%} & 39.38s  \\
    Sweep & \multicolumn{1}{c}{20.606} & \multicolumn{1}{c}{32.14\%} & \textless0.01s       & \multicolumn{1}{c}{0.269} & \multicolumn{1}{c}{32.78\%} & \multicolumn{1}{c}{\textless0.01s} \\
    \midrule
    
    POMO augx8 & \multicolumn{1}{c}{\makecell{15.754\\(±1.800)}} & 
    \multicolumn{1}{c}{\makecell{0.69\%\\(±0.649\%)}} & 0.01s & 
    \multicolumn{1}{c}{\makecell{21.154\\(±2.138)}} & \multicolumn{1}{c}{\makecell{4.87\%\\(±1.133\%)}} & \multicolumn{1}{c}{\textless0.01s}  \\
    
    BQ bs16 & \multicolumn{1}{c}{\makecell{15.806\\(±1.807)}} & 
    \multicolumn{1}{c}{\makecell{1.02\%\\(±1.015\%)}} & 0.11s & 
    \multicolumn{1}{c}{\makecell{20.362\\(±2.139)}} & 
    \multicolumn{1}{c}{\makecell{0.94\%\\(±0.950\%)}} & 0.56s  \\
    \midrule

    LEHD RRC 1000 & \multicolumn{1}{c}{\makecell{\textbf{15.629}\\\textbf{(±1.793)}}} & 
    \multicolumn{1}{c}{\makecell{\textbf{-0.11\%}\\\textbf{(±0.616\%)}}} & 1.01s     & 
    \multicolumn{1}{c}{\makecell{\textbf{20.095}\\\textbf{(±2.137)}}} & 
    \multicolumn{1}{c}{\makecell{\textbf{-0.38\%}\\\textbf{(±0.627\%)}}} & 19.69s  \\
    
    \midrule
    Neural Rewriter* & \multicolumn{1}{c}{16.100 } & \multicolumn{1}{c}{-} &   \multicolumn{1}{c|}{-}     & \multicolumn{1}{c}{-} & \multicolumn{1}{c}{-} &  \multicolumn{1}{c}{-}    \\
    DACT  & \multicolumn{1}{c}{16.202 } & \multicolumn{1}{c}{3.55\%} & 0.14s &   \multicolumn{1}{c}{23.230 } & \multicolumn{1}{c}{14.71\%} & 1.02s  \\
    \midrule
    
    DRHG (T=1000) & \multicolumn{1}{c}{\makecell{15.643 \\(±1.790)}} & \multicolumn{1}{c}{\makecell{-0.02\%\\(±0.647\%)}} & 2.37s       & \multicolumn{1}{c}{\makecell{\textit{20.233}\\\textit{(±2.133)}}} & 
    \multicolumn{1}{c}{\makecell{-0.16\%\\(±0.648\%)}} & 8.91s \\

    \toprule
      & \multicolumn{1}{c}{} & \multicolumn{1}{c}{\textbf{CVRP500}} & \multicolumn{1}{c|}{} 
    & \multicolumn{1}{c}{} & \multicolumn{1}{c}{\textbf{CVRP1K}} & \multicolumn{1}{c}{} \\
          
    \textbf{Method} & \multicolumn{1}{c}{\textbf{ Obj. }} & \multicolumn{1}{c}{\textbf{Gap}} & \textbf{Time}  & \multicolumn{1}{c}{\textbf{ Obj. }} & \multicolumn{1}{c}{\textbf{Gap}} & \textbf{Time} \\
    \midrule
    
    LKH3  & \multicolumn{1}{c}{37.229 } & \multicolumn{1}{c}{0.00\%} & 154.69s &   \multicolumn{1}{c}{37.090 } & \multicolumn{1}{c}{0.00\%} & 199.7s  \\
    HGS   & \multicolumn{1}{c}{36.561 } & \multicolumn{1}{c}{-1.79\%} & 112.50s &    \multicolumn{1}{c}{36.288 } & \multicolumn{1}{c}{-2.16\%} & 149.1s \\
    Sweep & \multicolumn{1}{c}{46.839 } & \multicolumn{1}{c}{25.81\%} & \textless0.01s       & \multicolumn{1}{c}{49.166 } & \multicolumn{1}{c}{32.56\%} & \textless0.1s  \\
    \midrule
    
    POMO augx8 & \multicolumn{1}{c}{\makecell{44.638\\(±3.112)}} & 
    \multicolumn{1}{c}{\makecell{19.90\%\\(±11.109\%)}} & 0.47s & 
    \multicolumn{1}{c}{84.898} & \multicolumn{1}{c}{128.89\%} & 4.7s  \\
    
    BQ bs16 & \multicolumn{1}{c}{\makecell{37.606\\(±4.216)}} & 
    \multicolumn{1}{c}{\makecell{1.01\%\\(±0.825\%)}} & 3.47s &
    \multicolumn{1}{c}{\makecell{38.147\\(±3.174)}} & \multicolumn{1}{c}{\makecell{2.88\%\\(±1.258\%)}} & 8.7s  \\
    \midrule
    
    GLOP-G(LKH3) & \multicolumn{1}{c}{-} & \multicolumn{1}{c}{-} &      \multicolumn{1}{c|}{-}  & 
    \multicolumn{1}{c}{\makecell{39.651\\(±3.779)}} & \multicolumn{1}{c}{\makecell{6.90\%\\(±2.013\%)}} & 0.8s  \\

    LEHD RRC 1000 & \multicolumn{1}{c}{\makecell{\textbf{37.100}\\\textbf{(±4.257)}}} & 
    \multicolumn{1}{c}{\makecell{\textbf{-0.35}\%\\\textbf{(±0.534\%)}}} & 56.25s      & 
    \multicolumn{1}{c}{\makecell{37.432\\(±3.237)}} & \multicolumn{1}{c}{\makecell{0.92\%\\(±0.874\%)}} & 202.5s \\
    
    SIL PRC1000* & \multicolumn{1}{c}{-} & \multicolumn{1}{c}{-} &       \multicolumn{1}{c|}{-}   & \multicolumn{1}{c}{\textbf{36.810 }} & \multicolumn{1}{c}{\textbf{-0.76\%}} & 78.8s \\
    \midrule
    
    DACT  & \multicolumn{1}{c}{46.393 } & \multicolumn{1}{c}{24.98\%} & 3.83s &    \multicolumn{1}{c}{} & \multicolumn{1}{c}{OOM} &     \\
    \midrule
    
    DRHG (T=1000) & \multicolumn{1}{c}{\makecell{37.718\\(±4.446)}} & 
    \multicolumn{1}{c}{\makecell{1.31\%\\(±1.035\%)}} & 12.60s &   
    \multicolumn{1}{c}{\makecell{39.932\\(±3.854)}} & \multicolumn{1}{c}{\makecell{7.66\%\\(±2.226\%)}} & 12.8s  \\
    
    \bottomrule
    \end{tabular}%

    \caption{Results on CVRP}
  \label{table-cvrp-all}%
\end{table*}%

\subsection{Experimental Results}

\begin{figure}[h]
\centering
\includegraphics[width=0.42\textwidth]{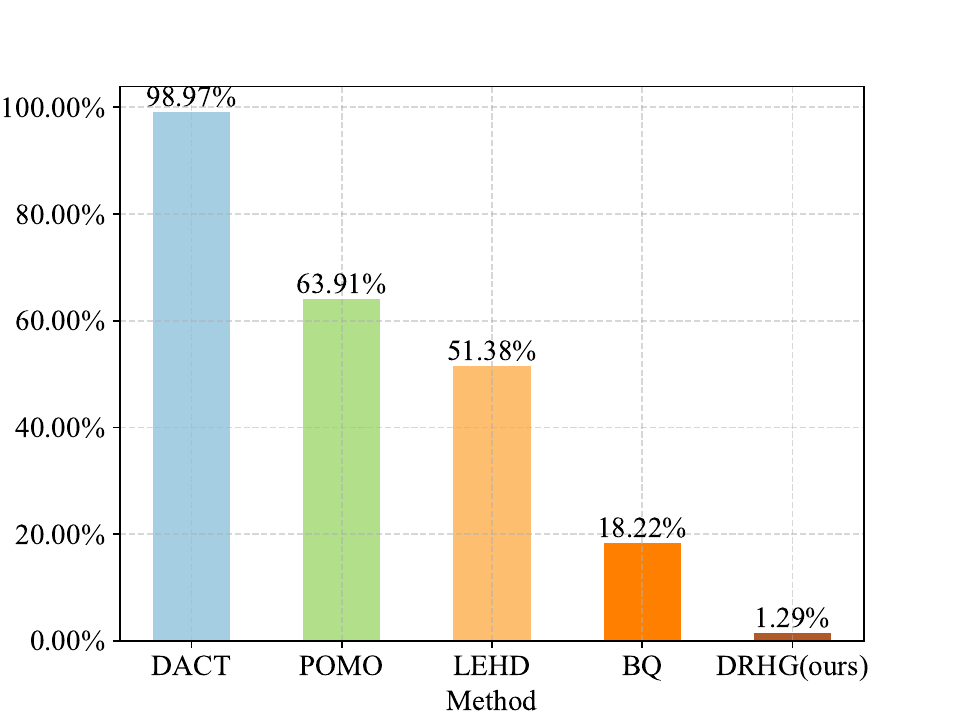} 
\caption{Non-optimal rate on 10k TSP100 instances}
\label{non-optimal}
\end{figure}

The main experimental results on uniformly distributed TSP instances are reported in Table \ref{table-tsp-all}. All results are reported in terms of per-instance solving time. The values in parentheses represent the variance. Rank-sum tests are conducted on POMO augx8, BQ bs16, GLOP (more revision) and LEHD RRC1000 to assess whether the path length and the gap of given methods differ significantly from those of our method. Except for the LEHD RRC1000 on TSP100, our DRHG demonstrates statistically significant differences ($p\textless0.05$) from other methods across all other test settings.

Particularly for TSP100, we track the number of non-optimal cases of the other representative NCO methods compared with our method, which is illustrated in Fig. \ref{non-optimal}. For TSPs with 100 to 500 nodes, we perform a comparative analysis of our method against other approaches under identical running time in Table 2, where $+$ means our method outperforms its competitor, and vice versa. Since the running time of one-shot constructive solvers cannot be adjusted, we allocate the same running time to our DRHG as that of its competitors.

The results demonstrate that our proposed DRHG method can achieve very good performance on instances of all sizes. 
Notably, on TSP100, our method yields non-optimal solutions in only 129 out of 10,000 cases, reducing the non-optimality ratio by an order of magnitude. On TSP200 and TSP500, our method reduces the gap by approximately one-third and outperforms its competitors given the same running time. With a small fine-tuning budget, our method outperforms all other methods on large-scale problems, including SIL \cite{luo2024SIL}, which is separately trained for each problem scale.

For CVRP (Table \ref{table-cvrp-all}), DRHG can also achieve pretty good performance. Our method outperforms the traditional heuristic method LKH3 on CVRP100 and 200. For CVRP200 and 500, our method outperforms most learning methods except for LEHD RRC1000, which, however, requires much more time. Overall, the performance of DRHG is slightly less dominant than that of TSP, but it is still promising.

Table \ref{tsplib} and Table \ref{cvrplib} show the test results on real-world TSPLib and CVRPLib instances with different sizes and distributions. The results show that our method is robust for different sizes and distributions. More results on TSPLib can be found in Appendix \ref{appendix-tsplib}. 

The ablation studies are presented in Appendix \ref{appendix-hyper-param}.

\begin{table}[htbp]
  \centering
    \begin{tabular}{l|c|c|c|c|c}
    \toprule
           & POMO  & BQ   & LEHD  & GLOP  & DRHG \\
    \multicolumn{1}{c|}{size} & augx8 & bs16  & R. 1K & more r. & T=1K \\
    \midrule
     \textless 100 & 0.79\% & 0.49\% & 0.48\% & 0.54\% & \textbf{0.48\%} \\
    100-200 & 2.42\% & 1.66\% & 0.20\% & 0.79\% & \textbf{0.15\%} \\
    200-500 & 13.41\% & 1.41\% & 0.38\% & 1.87\% & \textbf{0.36\%} \\
    500-1k & 31.68\% & 2.20\% & 1.21\% & 3.28\% & \textbf{0.26\%} \\
    \textgreater1k   & 63.71\% & 6.68\% & 4.14\% & 7.23\% & \textbf{2.09\%} \\
    \midrule
    all   & 26.41\% & 2.95\% & 1.59\% & 3.58\% & \textbf{0.95\%} \\
    \bottomrule
    \end{tabular}%
  \caption{Results on TSPLib}
  \label{tsplib}%
\end{table}%

\begin{table}[htbp]
  \centering
    \begin{tabular}{l|c|c|c|c|c}
    \toprule
          & POMO  & BQ   & LEHD  & GLOP  & DRHG \\
    \multicolumn{1}{c|}{size} & augx8 & bs16  & R. 1K & more r. & T=1K \\
    \midrule
    A & 4.97\% & 1.62\% & \textbf{0.75\%} & 26.18\% & 7.17\% \\
    B & 4.75\% & 4.06\% & \textbf{1.09\%} & 20.77\% & 5.55\% \\
    E & 11.40\% & 1.91\% & \textbf{0.58\%} & 18.25\% & 11.59\% \\
    F & 15.97\% & 7.36\% & \textbf{1.36\%} & 39.24\% & 33.18\% \\
    M & 4.86\% & 3.43\% & \textbf{1.43\%} & 22.60\% & 2.37\% \\
    P & 15.53\% & 2.01\% & \textbf{0.93\%} & 17.28\% & 7.53\% \\
    X & 21.68\% & 7.09\% & \textbf{3.69\%} & 18.48\% & 14.78\% \\
    \midrule
    All   & 15.45\% & 4.94\% & \textbf{2.36\%} & 20.10\% & 11.49\% \\
    \bottomrule
    \end{tabular}%
  \caption{Results on CVRPLib}
  \label{cvrplib}%
\end{table}%

\section{Conclusion, Limitation, and Future Work}
\paragraph{Conclusion} This paper has proposed a novel destroy-and-repair framework using hyper-graphs (DRHG) for routing problems. By leveraging the condensed hyper-graph formulation of the destroyed problem, we have reduced the burden of model learning and constrained the input size of the model to the scale of destruction. Extensive experiments comparing our model with other representative NCO methods on both synthetic and real-world instances have demonstrated the superiority of DRHG across different problem scales and distributions. 

\paragraph{Limitation and Future Work} The DRHG shows great performance on TSP, but our current design for CVRP has not fully realized the potential of DRHG. It could be interesting to ameliorate the implementation of DRHG and extend it to other routing problems. Furthermore, future work could explore more sophisticated destruction methods other than clustering destruction.

\section*{Acknowledgments}
This work was supported by the Research Grants Council of the Hong Kong Special Administrative Region, China (GRF Project No. CityU 11215622), the National Natural Science Foundation of China (Grant No. 62106096 and Grant No. 62476118), the Natural Science Foundation of Guangdong Province (Grant No. 2024A1515011759), the National Natural Science Foundation of Shenzhen (Grant No. JCYJ20220530113013031).

\newpage

\appendix

\section{Model Structure Details}\label{appendix-model}

\begin{figure}[htbp]
\centering
\includegraphics[width=0.35\textwidth]{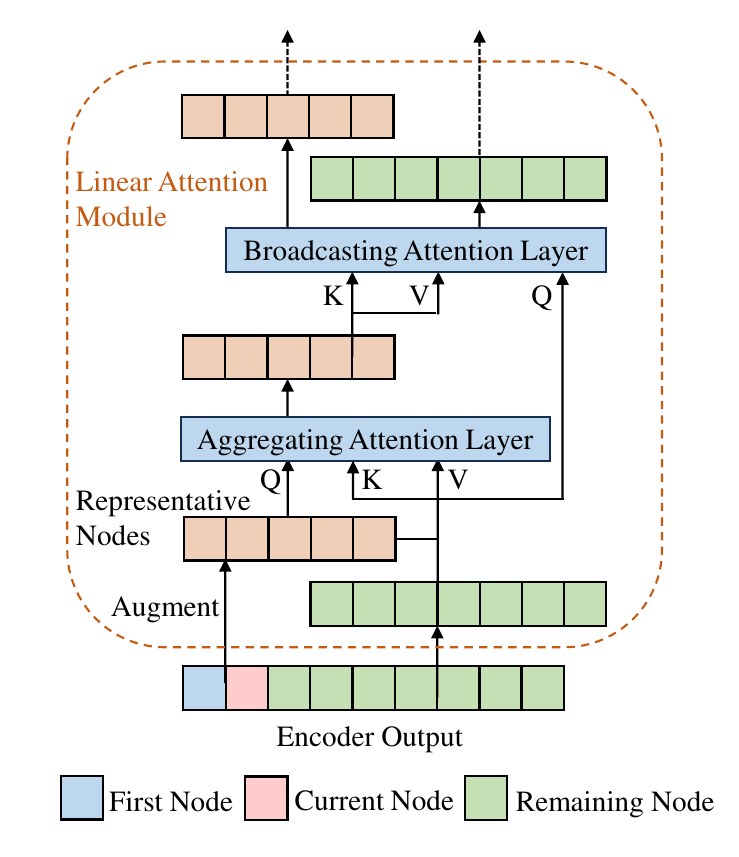} 
\caption{Linear attention module}
\label{linear attention}
\end{figure}

This section provides more details of the model structure, mainly about the decoder and the linear attention module, as shown in Fig. \ref{linear attention}. At each step $t$, the model takes the encoding of the first node, the current node, and the remaining unselected nodes for decoding. Denote the first and current nodes’ embeddings $h_f^{(0)}$ and $h_c^{(0)}$, respectively. We first augment them by $r_f$ and $r_c$ channels to obtain the virtual representative nodes: 
\begin{equation}
\Tilde{H}^0 = [reshape(h_f W_f), reshape(h_c W_c)],
\end{equation}
where $[\cdot, \cdot·]$ is the horizontal concatenation operator, $W_f \in \mathbb{R}^{d\times(d\times r_f)}$, $W_c \in \mathbb{R}^{d\times(d\times r_c)}$. Here the $reshape$ is to keep the embedding dimension of the representative node the same with the remaining unselected nodes, resulting $\Tilde{H}^0 \in \mathbb{R}^{(r_f + r_c)\times d}$, which means the number of virtual representative nodes equals the number of the channels. 

Then, we have $r = r_f + r_c $ virtual representative nodes embeddings $\Tilde{H}^{(0)} = \{h_j^{(0)} | j=1,2, ...,r\}$. Denote the remaining unselected nodes as $H_a^{(0)} = \{h_i^{(0)} | i = 1,2, ..., m-t\}$ at step $t$ for the hyper-graph of size $m$, the linear attention module first aggregates all information into representative nodes, then broadcasts the information to all nodes. 

The aggregating and broadcasting processes are realized by two different attention layers. Recall the formulation of classical attention mechanism: note the queries $X_Q \in \mathbb{R}^{q\times d} $, keys $X_K \in \mathbb{R}^{k\times d} $ and values $X_V \in \mathbb{R}^{v\times d}$ as inputs, the classical attention mechanism can be formulated as: 
\begin{equation}
\begin{split}
Attn&(X_Q, X_K, X_V) = \\
 &softmax(\frac{X_Q W_Q(X_K W_K)^\top}{\sqrt{d}})X_V W_V,
\end{split}
\end{equation}

Then, for the $l-th$ linear attention module, denote the input representative nodes' embeddings $\Tilde{H}^{(l-1)}$ and the remaining unvisited nodes’ embeddings $H_a^{(l-1)} = \{h_i^l, i=1, . . . , m\}$, we concat $\Tilde{H}^{(l-1)}$ and $H_a^{(l-1)}$ as $H_{all}^{(l-1)} = [\Tilde{H}^{(l-1)}, H_a^{(l-1)}]$. Then, the aggregating attention layer attends representative nodes to all nodes: 
\begin{equation}
\begin{split}
    Agg = Attn( & \Tilde{H}^{(l-1)}, H_{all}^{(l-1)}, H_{all}^{(l-1)}),
\end{split}
\end{equation}
and the broadcasting attention layer attends all nodes to the aggregations: 
\begin{equation}
Brd = Attn(H_{all}^{(l-1)}, Agg, Agg),
\end{equation}
where $Brd \in \mathbb{R}^{(r+m-t)\times d}$, which we can split into the representative nodes' embeddings $\Tilde{H}^{(l)}$ and the remaining unselected nodes’ embeddings $H_a^{(l)}$. 

Finally, after $L$ linear attention modules, we obtain the hidden representation $\Tilde{H}^{(L)}$ and and $H_a^{(L)}$. Then we take only the embeddings of remaining unselected nodes $H_a^{(L)}$ to calculate the probability for selecting the next node.

\section{Sample Size Alignment}\label{appendix-alignement}

\begin{figure}[htbp]
\centering
\includegraphics[width=0.4\textwidth]{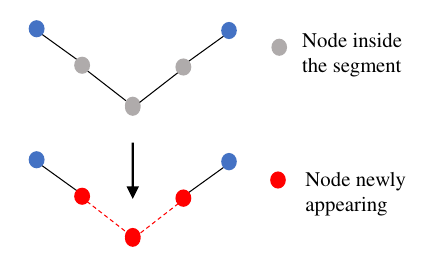} 
\caption{Node emergence in hyper-graph following additional node destruction}
\label{node_emerge}
\end{figure}

\begin{figure}[h]
\centering
\includegraphics[width=0.45\textwidth]{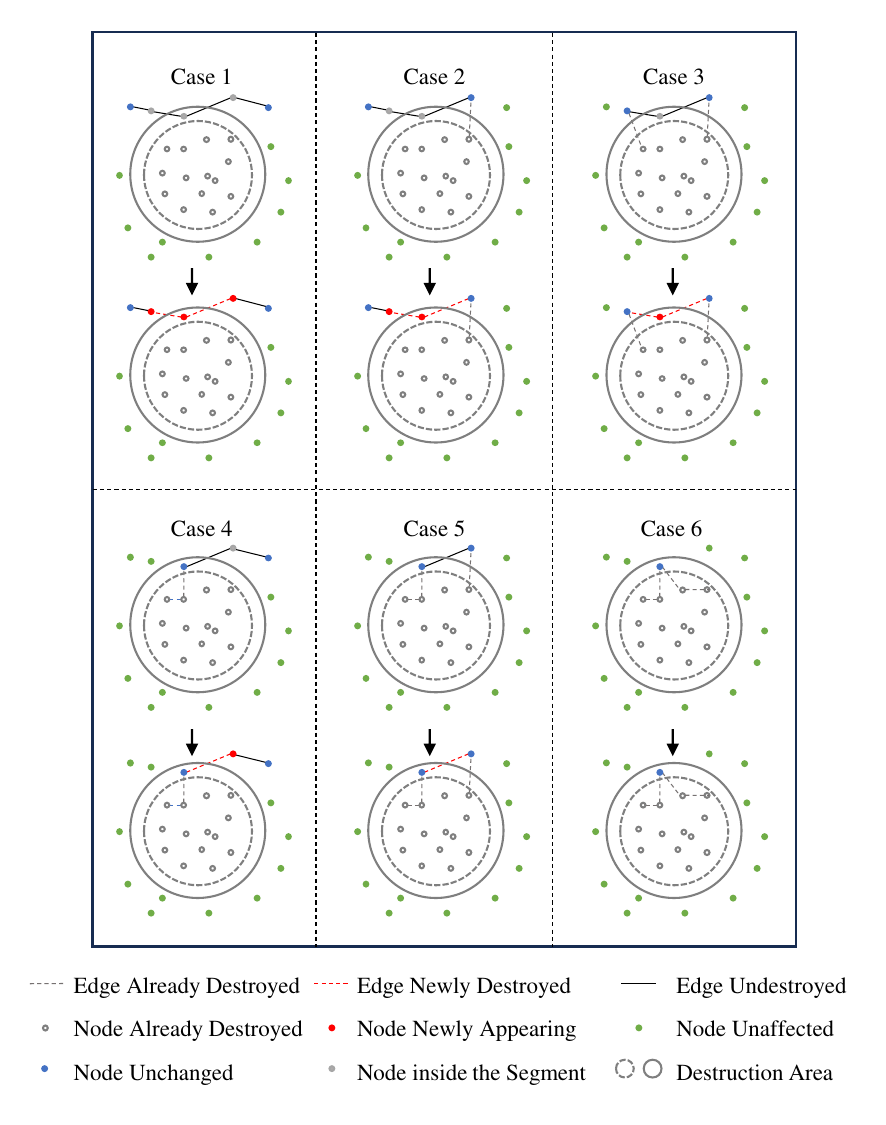} 
\caption{Different cases of node connections and corresponding results of newly appearing nodes when enlarging the destruction area}
\label{case-destruction}
\end{figure}

We employ the sample size alignment to make the hyper-graph size the same within a batch. The key concept is to precompute the size of the hyper-graph before actually implementing different cases of destruction. This process requires a predefined order of node destruction. In our clustering scenario, we begin with a central node, with nodes closer to the center being destroyed earlier. When an additional node is destroyed, the number of nodes that newly appear in the hyper-graph depends on whether the edges connecting the node to its neighbors (first-order and second-order) have already been destroyed. Fig. \ref{node_emerge} provides an example where the node targeted for destruction is connected with its four neighbors. After the node is destroyed, three new nodes emerge in the hyper-graph. We detail all cases of node connections and their corresponding results of node emergence when enlarging the destruction area in Fig. \ref{case-destruction}. 

There are six different cases and we indicate the newly appearing nodes by red color in each case. In case 1, destroying one more node generates two endpoint nodes and an isolated node. In case 2, that generates an endpoint node and an isolated node. In case 3, the newly destroyed node changes from a middle node to an isolated node. In case 4, the newly destroyed node becomes an endpoint node. In cases 5, the newly destroyed node is an endpoint node before and becomes an isolated nodes, but does not change the hyper-graph size. In case 6, the newly destroyed node has already been disconnected with its two neighbors before, therefore none of new nodes emerging in the hyper-graph due to the destruction. In summary, the number of newly appearing nodes equals to the number of relevant undestroyed edges minus one, except the case 6 where all edges have already been destroyed. 

We detail the algorithm of sample size alignment in Algorithm \ref{alignment}. First, we calculate the distance of all nodes to the center node and sort these node by the distance in ascending order. Then, for the node to destroy, if its neighbor is nearer to the center node, the neighbor will be destroyed before the node and the edge between them will be disconnect. Line 5 in the algorithm details the conditions where the node is still connected with its neighbors. After that, we can calculate the number of emerging nodes as line 6. Finally, by summing all emerging nodes in order, we can determine the hyper-graph size corresponding to a specific destruction scenario, and obtain the mask indicating which node should be destroyed for a target hyper-graph size.

\begin{algorithm}[]
    \caption{Sample Size Alignment}
	\begin{algorithmic}[1]

         \STATE {\bfseries Input:} 
         \newline The coordinates of a batch of VRP instances $X$,
         \newline the initial solution $S$, 
         \newline the cluster center $c=(x_c, y_c)$, 
         \newline the target reduced hyper-graph size $k$; 
         \STATE {\bfseries Notation:} 
         \newline $cumsum()$: the function to calculate the cumulative sum of a tensor along a dimension,
         \newline  $dist()$: the function to calculate the distance between two nodes,
         \newline $sort()$: sort a sequence of numbers in ascending order;
	    \STATE {\bfseries Output:} The mask $M$ indicating the node to destroy; 
        \STATE $D=dist(X, c)$, $\pi = sort(D)$; 
        \STATE For all nodes and their first-order neighbor $1A, 1B$ and the second-order neighbors $2A, 2B$ in the solutions, compare the distances to determine if they are connected with their neighbors before they are destroyed: 
        \newline \hspace*{3mm} $connect_{1A} = (D_{1A} \textgreater D)$,
        \newline \hspace*{3mm} $connect_{1B} = (D_{1B} \textgreater D)$,
        \newline \hspace*{3mm} $connect_{2A} = (D_{2A} \textgreater D)\ AND \ connect_{1A} $, 
        \newline \hspace*{3mm} $connect_{2B} = (D_{2B} \textgreater D)\ AND \ connect_{1B} $;
        \STATE Calculate the number of newly appearing nodes as illustrated in Fig. \ref{case-destruction}: 
        \newline \hspace*{3mm} $N = max(0, sum(connect_{1A} + connect_{1B} + connect_{2A}) + connect_{2B}) - 1)$; 
	    \STATE $H=cumsum(N)$ in the order of $\pi$; 
        \STATE $M = (H \leq k)$.
	\end{algorithmic}
	\label{alignment}
\end{algorithm}

\section{Additional Results of TSPLib}\label{appendix-tsplib}

The detailed results of TSPLib are shown in Table \ref{Detailed Tsplib result} and Table \ref{Detailed-TSPLib-continue}. Note that in two instances, rl11849 and usa13509, the LEHD \cite{luo2023lehd} needs about 80 hours and 120 hours to perform 1000 reconstruction steps. We stop the iteration at 24 hours as we observe that the LEHD has not made any progress for a long time. In most instances, our DRHG achieves the lowest optimality gap. The advantage of DRHG becomes more pronounced in hard instances of larger size or special distribution. 

\paragraph{large-size problem} In large-size problem, POMO \cite{kwon2020pomo} suffers the most from poor generalization ability, and its performance drops dramatically. BQ \cite{drakulic2023bq} and LEHD \cite{luo2023lehd} generalize better but still struggle on cases with thousands of nodes. When the problem size grows beyond 3,000, BQ fails to solve the problem due to out-of-memory, so to LEHD when the problem size comes to about 15,000. The DRHG succeeds in solving all instances and obtains solutions with optimality gap much lower than the others.

\paragraph{special distribution} Excepting the instances consisting of the real-world cities, the TSPlib also incorporates some instances of special distributions, such as the drilling problems (starting with 'd', 'pcb' and 'u'), and the rattled-grid problems (strat with 'rat'). On these distributions, DRHG also outperforms the other methods. 


\begin{table}[htbp]
  \centering
  \resizebox{0.9\columnwidth}{!}{

\begin{tabular}{ccccc}
\toprule
\textbf{Case} & \textbf{POMO-augx8} & \textbf{BQ-bs16} & \textbf{LEHD-RRC100} & \textbf{DRHG-T=1000} \\
\midrule
a280  & 12.62\% & 0.39\% & \textbf{0.30\%} & 0.34\% \\
\midrule
berlin52 & 0.04\% & \textbf{0.03\%} & \textbf{0.03\%} & \textbf{0.03\%} \\
\midrule
bier127 & 12.00\% & 0.68\% & \textbf{0.01\%} & \textbf{0.01\%} \\
\midrule
brd14051 & OOM   & OOM   & OOM   & \textbf{4.02\%} \\
\midrule
ch130 & 0.16\% & 0.13\% & \textbf{0.01\%} & \textbf{0.01\%} \\
\midrule
ch150 & 0.53\% & 0.39\% & \textbf{0.04\%} & \textbf{0.04\%} \\
\midrule
d1291 & 77.24\% & 5.97\% & 2.71\% & \textbf{2.09\%} \\
\midrule
d15112 & OOM   & OOM   & OOM   & \textbf{3.41\%} \\
\midrule
d1655 & 80.99\% & 9.67\% & 5.16\% & \textbf{1.57\%} \\
\midrule
d18512 & OOM   & OOM   & OOM   & \textbf{3.63\%} \\
\midrule
d198  & 19.89\% & 8.77\% & 0.71\% & \textbf{0.26\%} \\
\midrule
d2103 & 75.22\% & 15.36\% & \textbf{1.22\%} & 1.82\% \\
\midrule
d493  & 58.91\% & 8.40\% & 0.92\% & \textbf{0.31\%} \\
\midrule
d657  & 41.14\% & 1.34\% & 0.91\% & \textbf{0.21\%} \\
\midrule
eil101 & 1.84\% & 1.78\% & 1.78\% & \textbf{1.78\%} \\
\midrule
eil51 & 0.83\% & \textbf{0.67\%} & \textbf{0.67\%} & \textbf{0.67\%} \\
\midrule
eil76 & \textbf{1.18\%} & 1.24\% & 1.18\% & \textbf{1.18\%} \\
\midrule
fl1400 & 47.36\% & 11.60\% & 3.45\% & \textbf{1.43\%} \\
\midrule
fl1577 & 71.17\% & 14.63\% & 3.71\% & \textbf{3.08\%} \\
\midrule
fl3795 & 126.86\% & OOM   & 7.96\% & \textbf{4.61\%} \\
\midrule
fl417 & 18.51\% & 5.11\% & 2.87\% & \textbf{0.49\%} \\
\midrule
fnl4461 & OOM   & OOM   & 12.38\% & \textbf{1.20\%} \\
\midrule
gil262 & 2.99\% & 0.72\% & \textbf{0.33\%} & \textbf{0.33\%} \\
\midrule
kroA100 & 1.58\% & 0.02\% & \textbf{0.02\%} & \textbf{0.02\%} \\
\midrule
kroA150 & 1.01\% & 0.01\% & \textbf{0.00\%} & \textbf{0.00\%} \\
\midrule
kroA200 & 2.93\% & 0.50\% & \textbf{0.00\%} & \textbf{0.00\%} \\
\midrule
kroB100 & 0.93\% & 0.01\% & \textbf{-0.01\%} & \textbf{-0.01\%} \\
\midrule
kroB150 & 2.10\% & \textbf{-0.01\%} & \textbf{-0.01\%} & \textbf{-0.01\%} \\
\midrule
kroB200 & 2.04\% & 0.22\% & \textbf{0.01\%} & \textbf{0.01\%} \\
\midrule
kroC100 & 0.20\% & 0.01\% & \textbf{0.01\%} & \textbf{0.01\%} \\
\midrule
kroD100 & 0.80\% & 0.00\% & \textbf{0.00\%} & \textbf{0.00\%} \\
\midrule
kroE100 & 1.31\% & 0.07\% & \textbf{0.00\%} & 0.17\% \\
\midrule
lin105 & 1.31\% & 0.03\% & \textbf{0.03\%} & \textbf{0.03\%} \\
\midrule
lin318 & 10.29\% & 0.35\% & \textbf{0.03\%} & 0.30\% \\
\midrule
linhp318 & 12.11\% & 2.01\% & 1.74\% & \textbf{1.69\%} \\
\midrule
nrw1379 & 41.52\% & 3.34\% & 8.78\% & \textbf{1.41\%} \\
\midrule
p654  & 25.58\% & 4.44\% & 2.00\% & \textbf{0.03\%} \\
\midrule
pcb1173 & 45.85\% & 3.95\% & 3.40\% & \textbf{0.39\%} \\
\midrule
pcb3038 & 63.82\% & OOM   & 7.23\% & \textbf{1.01\%} \\
\midrule
pcb442 & 18.64\% & 0.95\% & \textbf{0.04\%} & 0.27\% \\
\midrule
pr1002 & 43.93\% & 2.94\% & 0.77\% & \textbf{0.67\%} \\
\midrule
pr107 & 0.90\% & 13.94\% & \textbf{0.00\%} & \textbf{0.00\%} \\
\midrule
pr124 & 0.37\% & 0.08\% & \textbf{0.00\%} & \textbf{0.00\%} \\
\midrule
pr136 & 0.87\% & \textbf{0.00\%} & \textbf{0.00\%} & \textbf{0.00\%} \\

\midrule
pr144 & 1.40\% & 0.19\% & 0.09\% & \textbf{0.00\%} \\
\midrule
pr152 & 0.99\% & 8.21\% & 0.27\% & \textbf{0.19\%} \\
\midrule
pr226 & 4.46\% & 0.13\% & \textbf{0.01\%} & \textbf{0.01\%} \\
\midrule
pr2392 & 69.78\% & 7.72\% & 5.31\% & \textbf{0.56\%} \\
\midrule
pr264 & 13.72\% & 0.27\% & \textbf{0.01\%} & \textbf{0.01\%} \\
\midrule
pr299 & 14.71\% & 1.62\% & 0.10\% & \textbf{0.02\%} \\
\midrule
pr439 & 21.55\% & 2.01\% & 0.33\% & \textbf{0.12\%} \\
\midrule
pr76  & 0.14\% & \textbf{0.00\%} & \textbf{0.00\%} & \textbf{0.00\%} \\
\midrule
rat195 & 8.15\% & 0.60\% & 0.61\% & \textbf{0.57\%} \\
\midrule
rat575 & 25.52\% & 0.84\% & 1.01\% & \textbf{0.36\%} \\
\midrule
rat783 & 33.54\% & 2.91\% & 1.28\% & \textbf{0.47\%} \\
\midrule
rat99 & 1.90\% & \textbf{0.68\%} & \textbf{0.68\%} & \textbf{0.68\%} \\
\midrule
rd100 & 0.01\% & \textbf{0.01\%} & \textbf{0.01\%} & \textbf{0.01\%} \\
\midrule
rd400 & 13.97\% & 0.32\% & \textbf{0.02\%} & 0.36\% \\

\bottomrule
\end{tabular}%
   }
    \caption{Detailed results of TSPLib}
  \label{Detailed Tsplib result}%
\end{table}%

\begin{table}[htbp]
  \centering
  \resizebox{0.9\columnwidth}{!}{
\begin{tabular}{ccccc}
\toprule
\textbf{Case} & \textbf{POMO-augx8} & \textbf{BQ-bs16} & \textbf{LEHD-RRC100} & \textbf{DRHG-T=1000} \\

\midrule
rl11849 & OOM   & OOM   & 21.43\% & \textbf{3.94\%} \\
\midrule
rl1304 & 67.70\% & 5.07\% & 1.96\% & \textbf{0.79\%} \\
\midrule
rl1323 & 68.69\% & 4.41\% & 1.71\% & \textbf{1.26\%} \\
\midrule
rl1889 & 80.00\% & 7.90\% & 2.90\% & \textbf{0.95\%} \\
\midrule
rl5915 & OOM   & OOM   & 11.21\% & \textbf{1.97\%} \\
\midrule
rl5934 & OOM   & OOM   & 11.11\% & \textbf{2.69\%} \\
\midrule
st70  & \textbf{0.31\%} & \textbf{0.31\%} & \textbf{0.31\%} & \textbf{0.31\%} \\
\midrule
ts225 & 4.72\% & \textbf{0.00\%} & \textbf{0.00\%} & \textbf{0.00\%} \\
\midrule
tsp225 & 6.72\% & -0.43\% & \textbf{-1.46\%} & \textbf{-1.46\%} \\
\midrule
u1060 & 53.50\% & 7.04\% & 2.80\% & \textbf{0.48\%} \\
\midrule
u1432 & 38.48\% & 2.70\% & 1.92\% & \textbf{0.49\%} \\
\midrule
u159  & 0.95\% & \textbf{-0.01\%} & \textbf{-0.01\%} & \textbf{-0.01\%} \\
\midrule
u1817 & 70.51\% & 6.12\% & 4.15\% & \textbf{2.05\%} \\
\midrule
u2152 & 74.08\% & 5.20\% & 4.90\% & \textbf{2.24\%} \\
\midrule
u2319 & 26.43\% & 1.33\% & 1.99\% & \textbf{0.30\%} \\
\midrule
u574  & 30.83\% & 2.09\% & 0.69\% & \textbf{0.24\%} \\
\midrule
u724  & 31.66\% & 1.57\% & 0.76\% & \textbf{0.27\%} \\
\midrule
usa13509 & OOM   & OOM   & 34.65\% & \textbf{11.82\%} \\
\midrule
vm1084 & 48.15\% & 5.93\% & 2.17\% & \textbf{0.14\%} \\
\midrule
vm1748 & 62.05\% & 6.04\% & 2.61\% & \textbf{0.54\%} \\
\bottomrule
\end{tabular}%

   }
    \caption{Detailed results of TSPLib (continued)}
  \label{Detailed-TSPLib-continue}%
\end{table}%

\begin{figure*}[!h]
\centering
\includegraphics[width=0.95\textwidth]{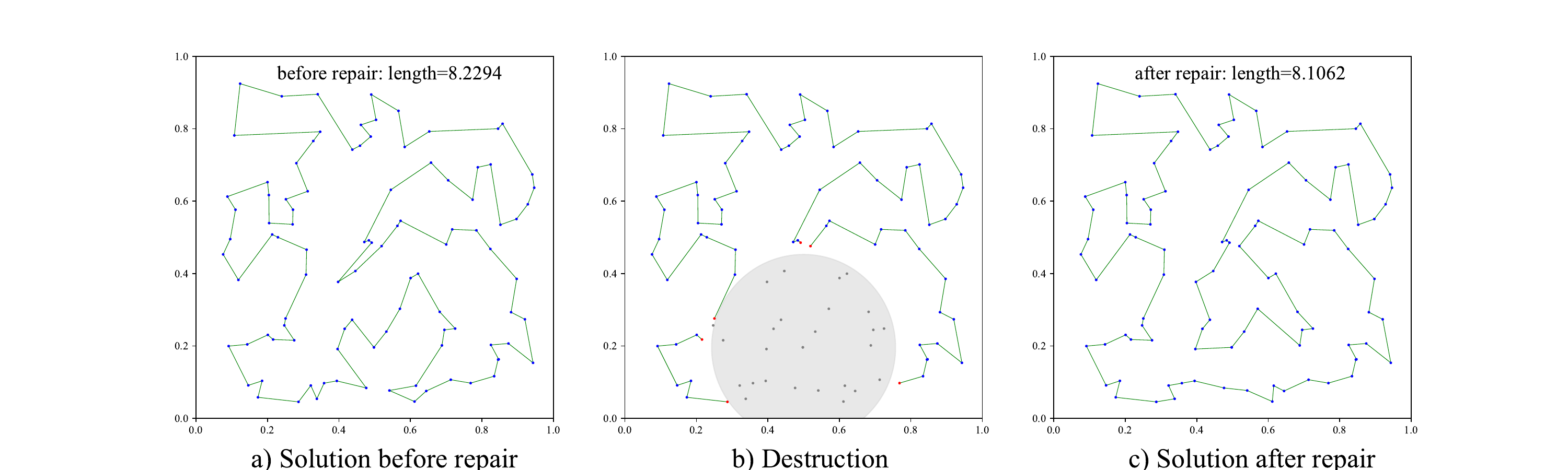} %
\caption{The demonstration of the destroy-and-repair process of a TSP instance}
\label{fig-DR_demo}
\end{figure*}

\section{Influence of Hyper-parameters}\label{appendix-hyper-param}
\paragraph{Effect of training sample size} The size of the training sample influences model performance by altering the distribution. A sample size that is too small may oversimplify the task, whereas a sample size that is too large may result in insufficient undestroyed segments for the model to effectively learn the repair process. We investigate this effect by training the model on TSP100, keeping all other settings constant. Table \ref{effect_train_size} shows the results of three settings where the sample size $\in [30, 70],\ [20, 80]$ and $[10, 90]$. The case where training sample size $\in [20, 80]$ performs the best.
\paragraph{Effect of destruction degree in the inference} Table \ref{destruction_size} illustrates the impact of destruction degree during inference on performance. A higher degree of destruction can be more efficient than a lower one, provided that the repair quality remains consistent. However, since the model is trained with no more than 100 nodes, repair quality diminishes when the destruction becomes too extensive. Destroying $k$ nodes with $k \in [20, 200]$ yields the best overall results.

\begin{table}[htbp]
  \centering
  \resizebox{0.8\columnwidth}{!}{

        \begin{tabular}{l|c|c|c}
        \toprule
        \multicolumn{1}{c|}{\multirow{2}[4]{*}{Gap}} & \multicolumn{3}{c}{Training sample size} \\
        \cmidrule{2-4}      & [30, 70] & [20, 80] & [10, 90] \\
        \midrule
        TSP100 & 0.0008\% & 0.0005\% & \textbf{0.0004\%} \\
        TSP200 & 0.014\% & \textbf{0.0098\%} & 0.0149\% \\
        TSP500 & 0.127\% & \textbf{0.113\%} & 0.121\% \\
        TSP1K & 0.268\% & \textbf{0.258\%} & 0.271\% \\
        TSP5K  & 1.47\% & \textbf{1.42\%} & 1.63\% \\
        TSP10K & 3.06\% & \textbf{2.85\%} & 3.384\% \\
        \bottomrule
        \end{tabular}%
   }
    \caption{Effect of training sample size}
  \label{effect_train_size}%
\end{table}%

\begin{table}[htbp]
  \centering
  \resizebox{0.9\columnwidth}{!}{
    \begin{tabular}{l|c|c|c|c}
    \toprule
    \multicolumn{1}{c|}{\multirow{2}[4]{*}{Gap}} & \multicolumn{4}{c}{Destruction size in the inference} \\
    \cmidrule{2-5}      & [20, 100] & [20, 200] & [20, 500] & [20,1000] \\
    \midrule
    TSP100 & 0.0005\% &       &       &  \\
    TSP200 & 0.0567\% & \textbf{0.0098\%} &       &  \\
    TSP500 & 0.239\% & 0.113\% & \textbf{0.111\%} &  \\
    TSP1K & 0.393\% & \textbf{0.258\%} & 0.309\% & 0.449\% \\
    TSP5K  & 1.70\% & \textbf{1.42\%} & 1.67\% & 2.05\% \\
    TSP10K & 3.34\% & \textbf{2.85\%} & 3.07\% & 3.46\% \\
    \bottomrule
    \end{tabular}%
   }
    \caption{Effect of destruction degree in the inference}
  \label{destruction_size}%
\end{table}%

\section{Destroy-and-repair Demonstration}\label{appendix-demo}

Fig. \ref{fig-DR_demo} demonstrates the destroy-and-repair of a TSP instance with $n=100$. Three segments are left after the destruction. The model changes how these segments are connected during the repair and makes the contour of the solution apparently different.

\end{document}